\documentclass[default,iicol]{sn-jnl}% Default with double column layout

\usepackage{float}
\usepackage{subcaption}
\usepackage{amsmath}
\usepackage{tabularx, booktabs} %% Load packages that you use

\newcolumntype{P}[1]{>{\centering\arraybackslash}m{#1}}
\jyear{2021}%

\theoremstyle{thmstyleone}%
\theoremstyle{thmstyletwo}%

\theoremstyle{thmstylethree}%

\begin{document}

\title[Article Title]{Virtually increasing the measurement frequency of LIDAR sensor utilizing a single RGB camera}

%%=============================================================%%
%% Prefix	-> \pfx{Dr}
%% GivenName	-> \fnm{Joergen W.}
%% Particle	-> \spfx{van der} -> surname prefix
%% FamilyName	-> \sur{Ploeg}
%% Suffix	-> \sfx{IV}
%% NatureName	-> \tanm{Poet Laureate} -> Title after name
%% Degrees	-> \dgr{MSc, PhD}
%% \author*[1,2]{\pfx{Dr} \fnm{Joergen W.} \spfx{van der} \sur{Ploeg} \sfx{IV} \tanm{Poet Laureate} 
%%                 \dgr{MSc, PhD}}\email{iauthor@gmail.com}
%%=============================================================%%

\author*[1,2]{\fnm{Zoltan} \sur{Rozsa}}\email{zoltan.rozsa@sztaki.hu}

\author[1,2]{\fnm{Tamas} \sur{Sziranyi}}\email{tamas.sziranyi@sztaki.hu}
%\equalcont{These authors contributed equally to this work.}

%\author[1,2]{\fnm{Third} \sur{Author}}\email{iiiauthor@gmail.com}
%\equalcont{These authors contributed equally to this work.}

\affil*[1]{\orgdiv{Machine Perception Research Laboratory}, \orgname{Institute for Computer Science and Control (SZTAKI), E\"otv\"os Lor\'and Research Network (ELKH)}, \orgaddress{\street{Kende u. 13-17.}, \city{Budapest}, \postcode{H-1111}, %\state{State},
\country{Hungary}}}

\affil[2]{\orgdiv{Faculty of Transportation Engineering and Vehicle Engineering}, \orgname{Budapest University of Technology and Economics (BME-KJK)}, \orgaddress{\street{M\H uegyetem rkp. 3.}, \city{Budapest}, \postcode{H-1111}, %\state{State},
\country{Hungary}}}

%\affil[3]{\orgdiv{Department}, \orgname{Organization}, \orgaddress{\street{Street}, \city{City}, \postcode{610101}, \state{State}, \country{Country}}}

%%==================================%%
%% sample for unstructured abstract %%
%%==================================%%

\abstract{The frame rates of most 3D LIDAR sensors used in intelligent vehicles are substantially lower than current cameras installed in the same vehicle. This research suggests using a mono camera to virtually enhance the frame rate of LIDARs, allowing the more frequent monitoring of dynamic objects in the surroundings that move quickly. As a first step, dynamic object candidates are identified and tracked in the camera frames. Following that, the LIDAR measurement points of these items are found by clustering in the frustums of 2D bounding boxes. Projecting these to the camera and tracking them to the next camera frame can be used to create 3D-2D correspondences between different timesteps. These correspondences between the last LIDAR frame and the actual camera frame are used to solve the PnP (Perspective-n-Point) problem. Finally, the estimated transformations are applied to the previously measured points to generate virtual measurements. With the proposed estimation, if the ego movement is known, not just static object position can be determined at timesteps where camera measurement is available, but positions of dynamic objects as well. We achieve state-of-the-art performance on large public datasets in terms of accuracy and similarity to real measurements.}


%%\pacs[JEL Classification]{D8, H51}

%%\pacs[MSC Classification]{35A01, 65L10, 65L12, 65L20, 65L70}

\maketitle

\section{Introduction}\label{sec1}
Both cameras and 3D LIDARs are elements of most sensor systems developed for autonomous cars to provide high-resolution color and direct depth information.
These LIDARs' usual frame rate is in the range of 5-20 Hz, while cameras work at least 30 FPS in most cases. We have to use the lowest measurement rate to get as high a horizontal (angular) resolution point cloud as possible from the LIDAR. However, if we are able to reconstruct the inter-frame point cloud of a LIDAR, we can get both high angular resolution and measurement frequency. Both high spatial and temporal resolution are essential in quick reaction. 

\begin{figure}[!h]
  \vspace{-0.2cm}
  \centering
  {\includegraphics[width=3in]{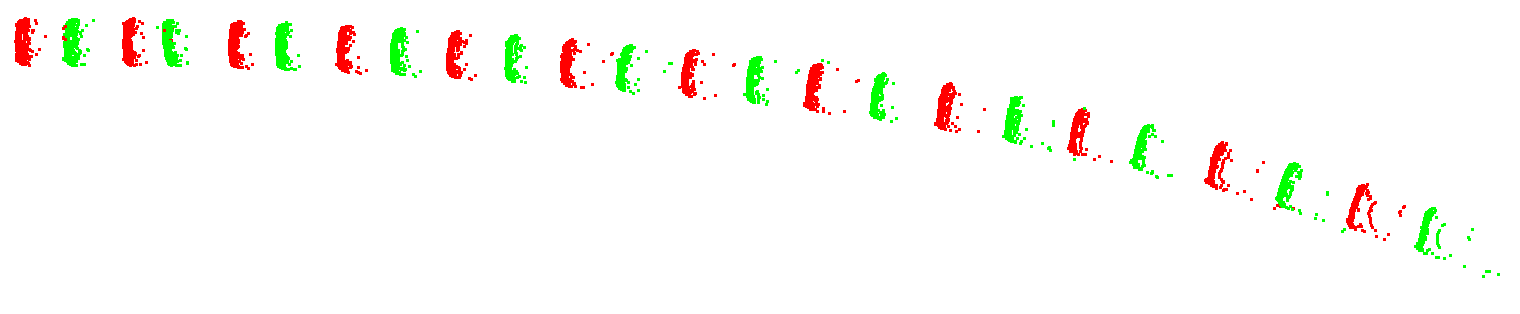}}
  \caption{Illustration of upsampled object trajectory by the proposed method in the coordinate system of the first LIDAR frame. Not every frame is illustrated for better visualization. Colormap: Red - real measurements; Green - virtually generated measurements.}%; Black - static points of the first LIDAR frame.}
  \label{fig:ups}
  \vspace{-0.1cm}
\end{figure}

%Spatial resolution is important in the immediate detection of far objects.
Spatial resolution decreases with the distance from the sensor. That is why, as soon as possible detection of distant objects is enabled only by the highest obtainable point cloud resolution.

A high temporal resolution has the same importance in the time of the first detection. If an object appears inter-frames, the detection delay should be as small as possible. Also, it is necessary to have the second measurement with the minimum time delays, as only it can provide information about the velocity of the tracked object \cite{9200682}. Later on, monitoring dynamic objects' trajectory as frequently as possible is also essential as vehicles are capable of quick direction change, high acceleration, or deceleration (which could result in a dangerous situation).

\begin{figure*}[h]
\centering
\subcaptionbox{Corresponding camera frames.\label{fig:det1}}
{\includegraphics[width=1.0\textwidth]{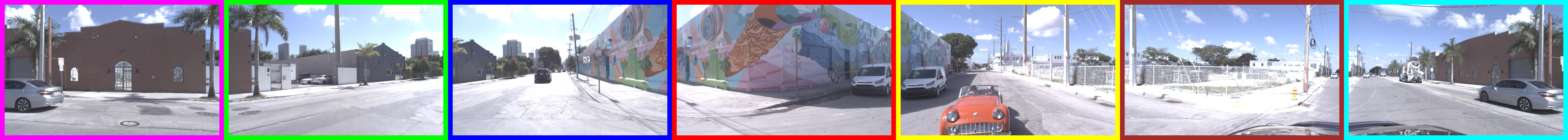}}
\subcaptionbox{Generated virtual LIDAR frame (estimations from different cameras indicated with different colors) and ground truth frame (colored by Black).\label{fig:det2}}
{\includegraphics[width=1.0\textwidth]{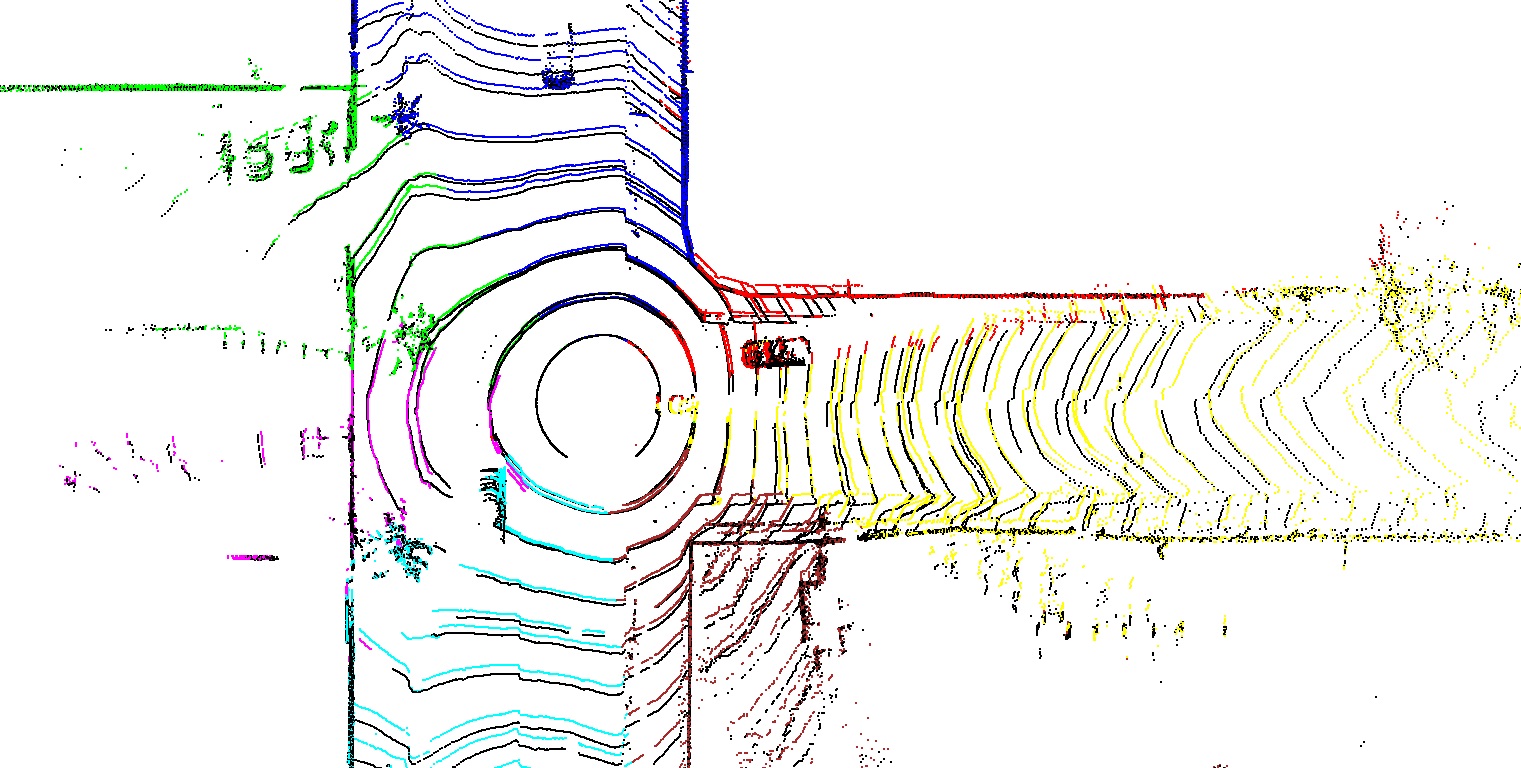}}
\caption{Example of using multiple cameras (from the Argoverse dataset \cite{DBLP:journals/corr/abs-1911-02620}) pointing at different directions to generate whole circular LIDAR frame. Images and LIDAR cloud parts generated based on the given view are indicated with the same color: Magenta - side left; Green - front left; Blue - front center; Red - front right; Yellow - side right; Brown - rear right; Cyan - rear left }\label{fig:whole}
\end{figure*}

In this paper, we propose a novel method to generate virtual LIDAR frames based on camera measurements to monitor dynamic objects in 3D more frequently than it is allowed by the LIDAR's sampling rate. A temporally up-sampled vehicle point cloud and trajectory are illustrated Fig. \ref{fig:ups}.
The precondition of applying the method is the presence of at least one calibrated camera-LIDAR pair, with a camera higher measurement frequency than the LIDAR. In this case, the LIDAR measurement can be temporally up-sampled in the common field of view. If there are multiple cameras available with different views, it is possible to temporally up-sample the whole circular LIDAR frame in $360^{o}$ (Fig. \ref{fig:whole}). Stereo arrangement of cameras can be applied, but not necessary, and we do not deal with this scenario as stereo matching can be computationally intensive.

Our framework estimates transformations describing movements of dynamic objects. This estimation is realized by object and point level tracking on a high-frequency camera and solving the PnP problem between $t-1$ (LIDAR) and $t$ (camera) time instances. The estimated transformations are used to generate interframe point clouds. %As a first step, dynamic object candidates are identified and tracked in the camera frames. 

Minimum delay reaction to the movement of surrounding agents is essential and intended to solve by forecasting in most of the cases \cite{ijcv}. However, the accuracy of these methods is limited by the temporal availability of position measurements of the agents. Our proposal greatly enhances this to solve the temporal upsampling problem and provide more recent agents positions than the hardware enables. Spatial upsampling \cite{hu2020PENet} and preceding inter-frame estimation \cite{spinet} has been addressed earlier. However, the temporal upsampling with the generation of actual interframes has not been solved yet, despite the extensive application areas of LIDARs (Advanced Driver Assistance Systems, automated transportation, mobile robotics, surveillance, etc.). This fact is explained by that the accurate reconstruction of dynamic objects (in terms of position and similarity to real measurements) in real-time is challenging.

\begin{figure*}[!h]
  \vspace{-0.2cm}
  \centering
  {\includegraphics[width=\textwidth]{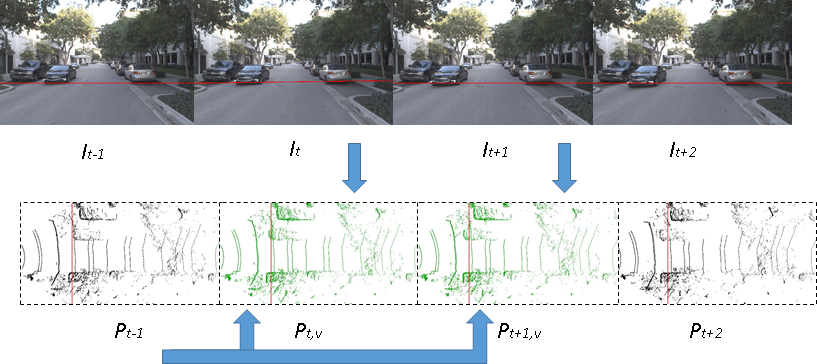}}
  \caption{Illustration of the method's goal and workflow. We generate virtual LIDAR measurements based on the last LIDAR measurement and current camera frame by utilizing the higher camera sampling rate. The above data is from the Argoverse dataset (sampling frequency: LIDAR - 10 Hz, camera - 30 Hz) \cite{DBLP:journals/corr/abs-1911-02620}.% For the illustration only every 3rd frame is used. 
  Colormap of the point clouds: Black - real measurements; Green - virtual measurements generated by the proposed method. Red line (horizontal on camera image and vertical on LIDAR point clouds) indicates the position of an approaching car in the opposing lane at $t-1$.}%; Black - static points of the first LIDAR frame.}
  \label{fig:timeline}
  \vspace{-0.1cm}
\end{figure*}

Preliminary results about the method were introduced in \cite{ICIAP_ZR_TSZ}. An illustration of the goal and results generated by the method is visible in Fig. \ref{fig:timeline}. $I$ and $P$ indicate camera image and LIDAR point cloud respectively, $v$ index means virtually generated, the timestamps go from $t-1$ to $t+2$ in the illustration. In the rest of the paper, we will explain how the method works on the basis of $t-1$ and $t$ timestamps, as to others, it can be easily extended.

\subsection{Contribution}
%The main contribution of the paper is the proposal of a novel methodology to temporally up-sample LIDAR point clouds based on camera measurements.
%The advantages of the method compared to others currently available with similar purpose):
The most main contribution of the paper is that it allows us to get 3D information about the position of dynamic objects with a higher frequency than the measurement frequency of the LIDAR sensor. The benefits of our method are listed below:
\begin{itemize}
    \item Processing steps which are required by high-level environment perception algorithms (change detection \cite{s20072068}, SLAM, online-fusion, etc. \cite{BENEDEK2021103193}) like fusion, object detection, or tracking are already included in our pipeline (with a real-time run). In this way, these processing steps do not have to be executed again, as would be the case in any deep-learning-based Pseudo-LIDAR algorithms.
    \item The proposed method uses only one previous LIDAR measurement to generate a virtual one (other methods - like \cite{Weng2020_SPF2} or \cite{9652466} - use five or more).
    \item Our proposed method requires no learning because it is based on geometry.
    \item The method can be applied to LIDAR point clouds with different characteristics and generate results with the same characteristics. This fact came from the working principle (as transformations are applied at object-level) and was proved by evaluating on different databases (Section \ref{sec:test}) with different LIDAR sensors.
   %\item The method can process a higher range in real-time than other currently available prediction methods.
    
\end{itemize}

%\vfill
\subsection{Outline of the Paper}

The paper is organized as follows: Section \ref{sec:relw} surveys the literature about the related works. Section \ref{sec:propm} describes the proposed method in detail. Section \ref{sec:test} shows our test results and Section \ref{sec:disc} discuss them.  Finally, Section \ref{sec:conclusion} draws the conclusions.

\section{Related Work}
\label{sec:relw}

Two subsections are discussed in this section. First, camera-based depth estimation and completion literature is introduced, which relates to spatial upsampling. Then, methods for virtual LIDAR frame generation are investigated. These either generate intermediate frames but cannot be applied for the given (temporal upsampling) purpose because of their preconditions or they can be applied (but not designed for that purpose). Later, these are considered in comparison, as there are no available approaches for temporal upsampling. %Then the most significant methods (in our point of view) are introduced, which deal with point cloud upsampling (but with different preconditions).

\subsection{Depth Prediction and Completion}
A partial solution can be provided to the temporal up-sampling problem using only camera depth estimation methods (the resulted point cloud will not have LIDAR characteristics). In the literature direct depth estimation algorithms are available (like \cite{Qiao_2021_CVPR} or \cite{Fu_2018_CVPR}). Alternatively, algorithms aiming the reconstruction of static scene could be used together with methods (like \cite{7298997}, \cite{6909599} and \cite{8708251}) which are designed for mono camera-based trajectory estimation. In this way, they can reconstruct moving objects' paths to generate a 3D point cloud for every time instance (when camera measurement is available). 
%(Stereo camera-based depth estimation has a similar frequency as LIDAR measurements, that is why they are not suitable for temporal up-scaling. \cite{1467526}). 
Naturally, the above methods have their disadvantages:
\begin{itemize}
    \item They are also mostly neural network-based, so training is required before inference in most cases.
    \item Scaling has to be solved in the case of the mono camera.
    \item The estimated point clouds do not have LIDAR characteristics, so algorithms designed for LIDARs cannot be easily adapted for further processing.
\end{itemize}
In Section \ref{sec:test} we provide comparisons to these in terms of accuracy.
Methods like \cite{wang2019pseudo} and \cite{wang2019pseudo2} uses stereo camera pair to generate so-called pseudo LIDAR measurements. They are similar to our virtual measurements, but they are different in the following:
\begin{itemize}
\item in purpose: These methods increase detection performance in a stereo point cloud. We utilize mono camera-based detections (with a higher detection rate than point clouds generally have) and help any further LIDAR processing.
\item in frequency: Stereo image pair processing will not enable a higher frame rate that LIDAR already has, as it has similar frequency \cite{1467526}.
\item in characteristics: They basically leave stereo matching step and compute point cloud directly from stereo image pair. The result will not be similar to LIDAR measurements.
\end{itemize}

As we mentioned earlier, existing LIDAR sensors have a much lower resolution than cameras in terms of frequency and spatial frequency. 

Increasing spatial resolution of the LIDAR sensor can be realized in different ways, e.g., by registering frames \cite{8283563} or utilizing camera images. In the latter case (spatial up-sampling by the camera), range images generated from LIDAR measurements to match the resolution of camera images. LIDARs operate in the range of several hundreds of thousands of data points, but cameras have millions of pixels. This problem has diverse literature. Neural network-based methods (\cite{9440471}, \cite{hu2020PENet}) offer the state-of-the-art performance for the depth completion \cite{ijcv2}, in the KITTI \cite{Geiger2013IJRR} dataset of this task \cite{Uhrig2017THREEDV}. However, there are also conventional methods available using semantic-based upsampling \cite{10.1007/978-3-319-45886-1_4} or bilateral filtering \cite{inproceedingsus}.

The ratio of sampling frequencies of camera and LIDAR frames (complete $360^{o}$ degree rotation) is about 3 or more in modern systems (e.g.: in \cite{DBLP:journals/corr/abs-1911-02620} $\frac{f_{C}}{f_{L}} = \frac{30}{10} $). Still, there are no methods available to solve the temporal upsampling problem up to now. 

\subsection{Virtual LIDAR Frame Generation}
In this subsection, alternative ways are discussed for generating 3D point clouds for a time instance without LIDAR measurement. All of these operate using only LIDAR data. The first one of the two strategies discussed here is the prediction of future point clouds based on previous ones.

Nowadays, sequential point cloud forecasting is a hot research topic. Papers like \cite{Weng2020_SPF2}, \cite{9389722} \cite{9652466} or \cite{ijcv} offer possibilities of forecasting future LIDAR point cloud usually from 5 previous frames. Relying on so many frames and only past information are serious drawbacks. The system learns to forecast an object's state information from these numerous frames. If a new object appears (only visible in the last few frames), its pose will likely be erroneously forecasted. If it was perceptible only in the last frame, it would not answer even the most basic question of whether it is a static or dynamic object. With information from the current timestamp (what we gain from the camera), these can be estimated. Even high accelerations or direction changes do not pose a threat. 
Finally, these prediction methods are deep-learning-based approaches that imply that training is required to operate and re-training for point clouds of different resolutions and characteristics. %Finally, they are not yet applicable in practice as forecasting pipelines operate only in close range quasi-real-time only yet \cite{8818349}.

The most similar works are related to the temporal interpolation of LIDAR clouds. In \cite{articleplin}, \cite{9352507}, \cite{Lu2020_PointINet} and \cite{spinet}, the authors solve the frequency mismatching or synchronization problem of cameras and LIDARs. These interpolation methods require a 'future' LIDAR frame (in our point of view) to generate the inter-frame point cloud. They can be applied only in an "offline" manner, so they cannot be an alternative for our proposal. In practice, the camera frame with the closest timestamp to the LIDAR frame's timestamp is used most often as a matching frame. We use this assumption as well without dealing with frequency mismatching. If the initial data has a serious synchronization problem ($\frac{f_{C}}{f_{L}} <=1$, which is not typical), these methods can be used as a preprocessing for preceding frames. Mismatching can be caused by high-speed ego-movement, too, because the first data point of the LIDAR can be measured from a substantially different viewpoint than the last measurement point. However, methods like \cite{deskew} offer a solution for that problem too. Applying high-frequency localization equipment enables the transformation of the LIDAR points to the same coordinate system.

To summarize, every alternative of our proposal introduced in this subsection has serious drawbacks compared to our method using camera and LIDAR data. Besides, we will show in Section \ref{sec:test} that our proposed method also outperforms these in the comparison in real-time.

%\begin{figure}[!h]
%\centering
%\subcaptionbox{Point cloud of $T_{t-1}$ (detected dynamic objects are colored with blue) \label{fig:t_1}}
%{\includegraphics[width=0.51\columnwidth]{Figures/frame_t_1_2.png}}
%\subcaptionbox{Ground truth point cloud of $T_{t}$ (detected dynamic objects are colored with red) \label{fig:gt}}
%{\includegraphics[width=0.49\columnwidth]{Figures/frame_t_2.png}}
%\subcaptionbox{Estimated point cloud of $T_{t}$ (dynamic objects are colored with green) \label{fig:es}}
%{\includegraphics[width=0.49\columnwidth]{Figures/frame_t_es_2.png}}
%\subcaptionbox{Yellow highlighted vehicle at $T_{t-1}$ (blue) and $T_{t}$ (red) \label{fig:gt_t}}
%{\includegraphics[width=0.49\columnwidth]{Figures/t_1_gt_t_gt.png}}
%\subcaptionbox{Yellow highlighted vehicle's ground truth (red) and estimated (green) position at $T_{t}$ \label{fig:gt_es_t}}
%{\includegraphics[width=0.49\columnwidth]{Figures/t_es_gt.png}}
%\caption{Illustration of point cloud estimation of the proposed method. (a) subfigure shows the last LIDAR frame ($P_{t-1}$), (b) shows the ground truth and (c) our estimation of the point cloud at $T_{t}$. %of the first subfigure (static objects are transformed based on known ego-pose). 
%Subfigure (d) highlights ground truth points of a vehicle at $T_{t-1}$ and $T_{t}$. While (e) visualizes together the ground truth point cloud and our estimation.}\label{fig:pc_est}
%\end{figure}

\section{The Proposed Method}
\label{sec:propm}
As step zero, the intrinsic calibration \cite{888718} of the camera should be executed together with the LIDAR-camera calibration \cite{8593660}. Intrinsic matrix will be indicated as \textbf{$K$} and transformation matrix from the LIDAR coordinate system to the camera coordinate system as \textbf{$T_{L,C}$}.

The main steps of the method are the following: 

First, we detect the dynamic object candidates on images and determine their 3D points in the LIDAR data.
The detection is done for every frame, and objects are also tracked. In this way in timestamp $t$, we know the previous positions of these objects both in image and LIDAR coordinates in timestamp $t-1$. (We know the poses for preceding timestamps as well, but we do not use them.)
Besides, we also track 2D projections of 3D object points in camera images. Utilizing them, we can compute virtual camera motions assuming static objects. As real camera motion is known, object movements can be expressed from the virtual one. Finally, using the real camera motion and the estimated object movements, we can transform the points of the last LIDAR frame's static and dynamic objects respectively with the calculated transformations to the current timestamp.

\textbf{Note:} As ego-motion assumed to be known from localization sensors \cite{deskew} - or could be determined solely from the camera by methods like \cite{orbslam} -, static scene is ignored in most of the pipeline. As we utilize ego-motion only relative to the previous LIDAR frame, there is no accumulation of localization error, so the chosen method for self-localization does not significantly influence our final result.

\subsection{Object Detection and Tracking on Camera Images}
\label{det_and_track}
In our tests, only vehicles are detected and tracked as they are the most frequent dynamic participants of road traffic. Pedestrians usually can be well monitored at the available LIDAR frequencies. However, as the detector can be replaced in the pipeline with any pretrained one, the detection and tracking can be used for any category as needed. Yolo\_v2 \cite{8100173} vehicle detector was used in our experiments. 
For the tracking on the object level, we applied the Hungarian algorithm \cite{Miller97optimizingmurty's} to solve the assignment problem.

\begin{figure}[!h]
\centering
\subcaptionbox{$I_{t-1}$ \label{fig:det1}}
{\includegraphics[width=1.0\columnwidth]{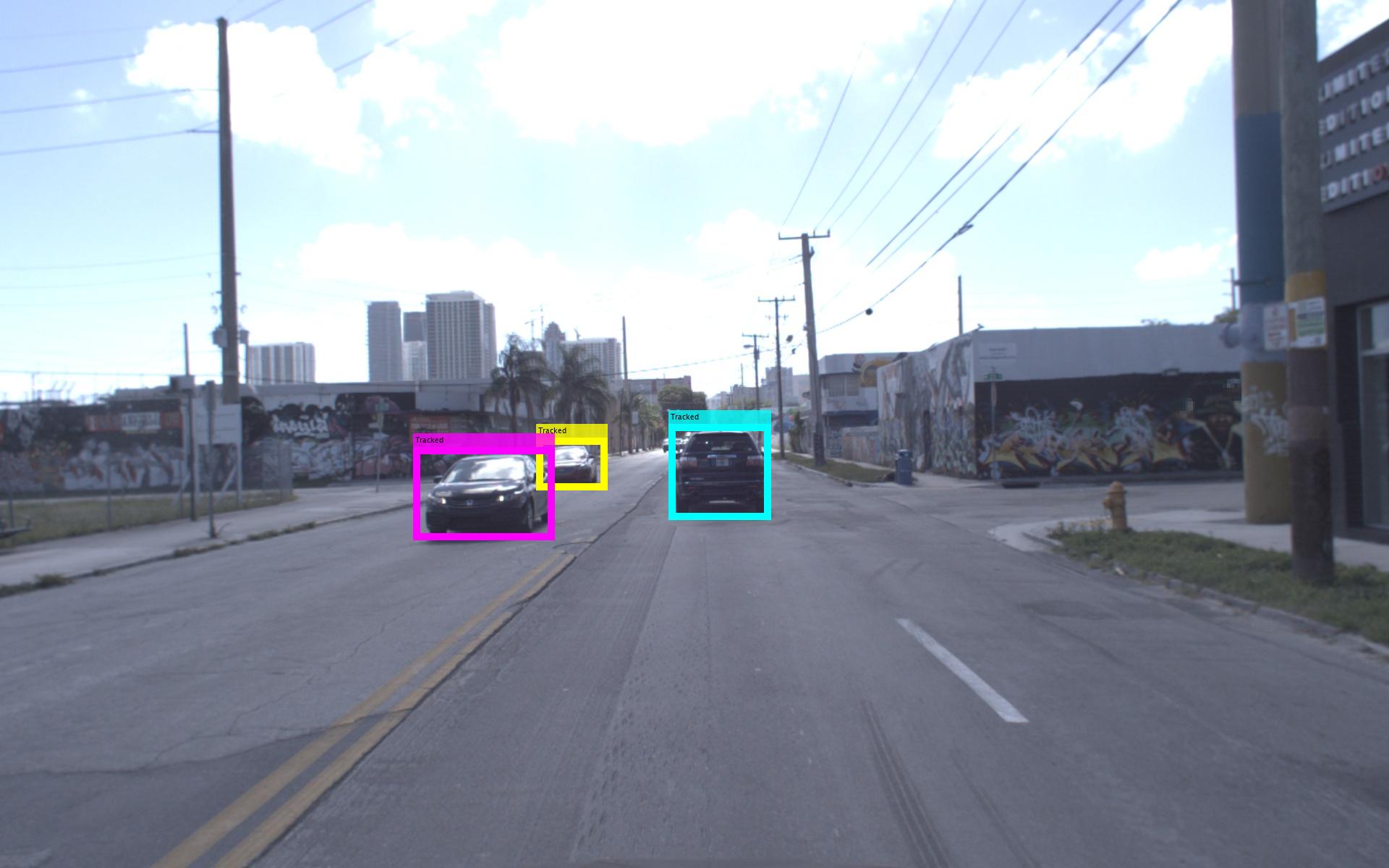}}
\subcaptionbox{$I_{t}$ \label{fig:det2}}
{\includegraphics[width=1.0\columnwidth]{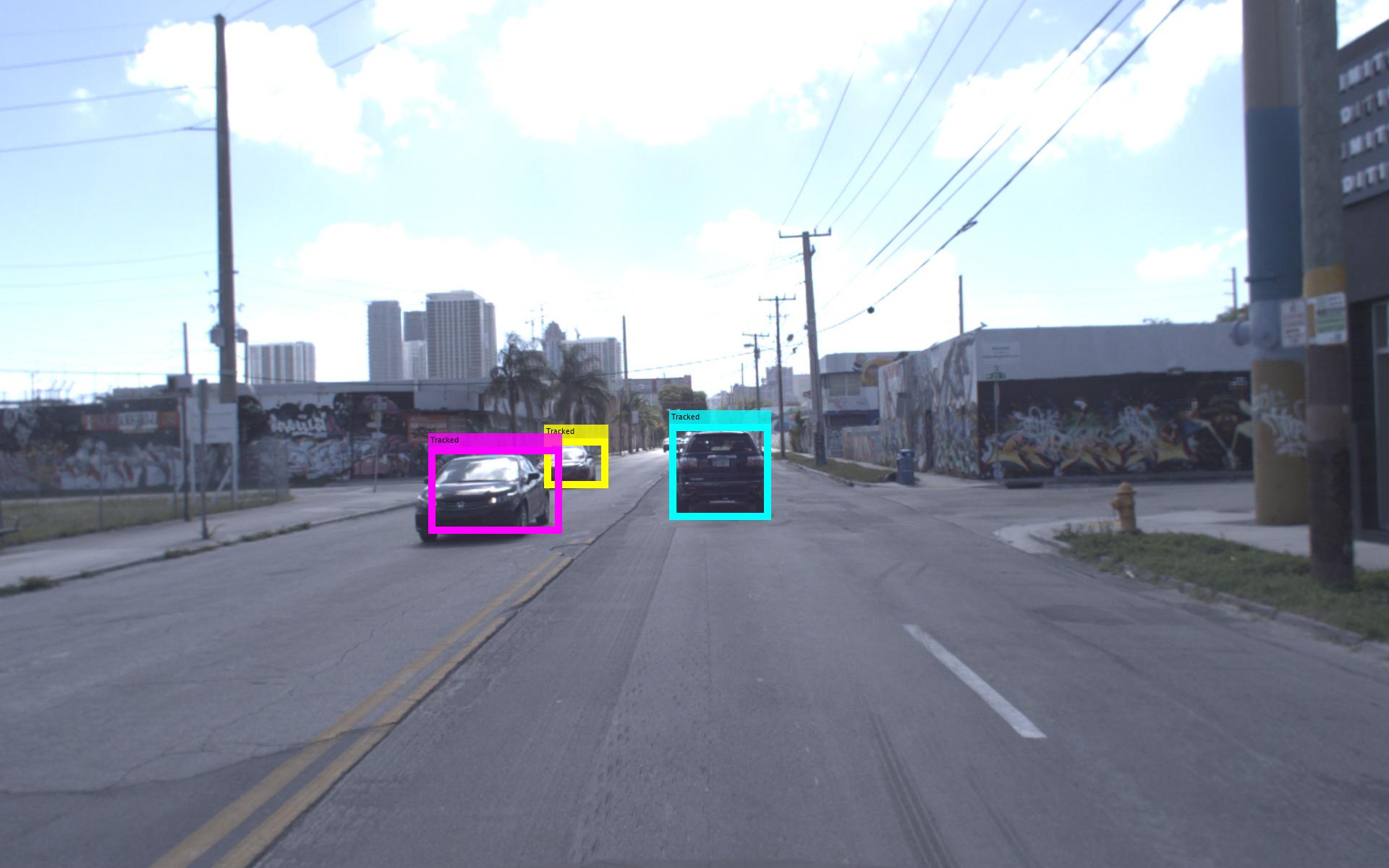}}
\caption{Example detections and tracks (indicated with different colors) on the Argoverse dataset \cite{DBLP:journals/corr/abs-1911-02620}.}\label{fig:track}
\end{figure}

Example detections and tracks can be seen in Fig \ref{fig:track}.

The detection and tracking steps are executed on each camera frame.
We estimate the movement of all of the tracked vehicles as interpreting their resulted transformation matrix it will be revealed whether they are static or dynamic. We do this instead of utilizing prior knowledge (from the preceding frames) in order to avoid problematic cases of sudden departing or stopping.

\subsection{Defining Object Points and Tracking Them}
We will use object points on $I_{t}$ to determine object transformations. Instead of searching for keypoints in the camera images, we will use object points for which we already had depth values at $t-1$ timestamp. We define salient points at $t-1$ timestamps as the projections of the LIDAR point cloud of the corresponding objects.

\begin{figure}[!h]
  \vspace{-0.2cm}
  \centering
  {\includegraphics[width=3in]{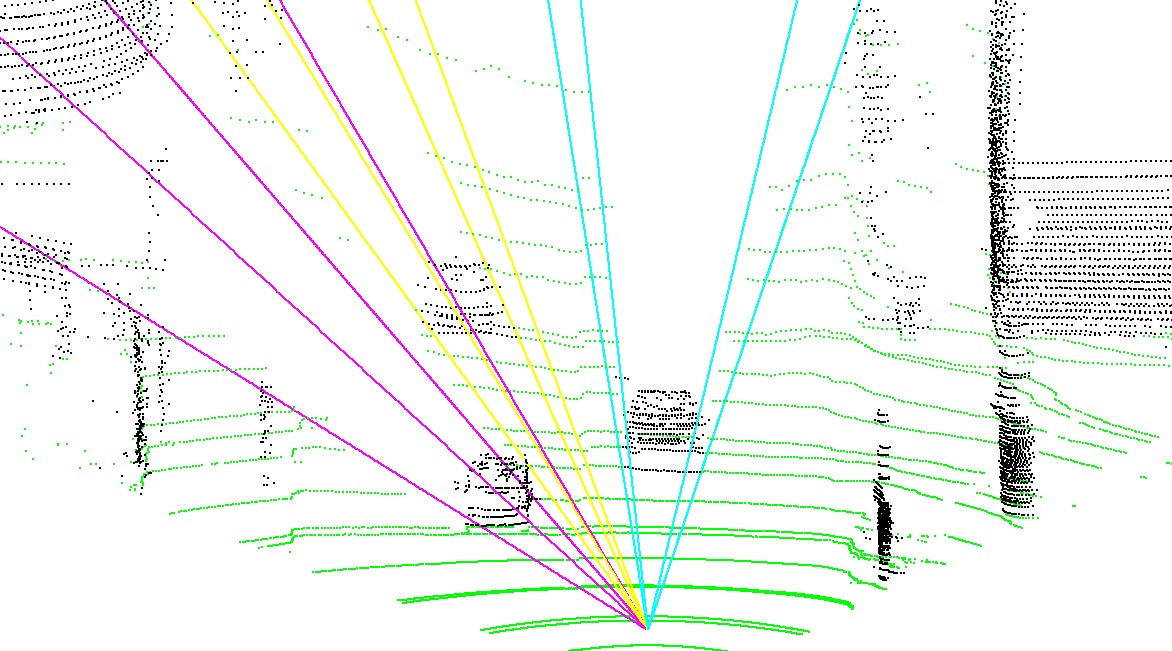}}
  \caption{Segmented ground points are colored with green. These are in the frustums of vehicle bounding boxes (resulted in 2D detections). Frustum colors indicates the same objects as in Fig. \ref{fig:track}}
  \label{fig:ground}
  \vspace{-0.1cm}
\end{figure}

To do this, first, we have to establish correspondences between 2D bounding boxes on $I_{t-1}$ and 3D points in the $P_{t-1}$. For each object, we select 3D measurement points that project to the 2D bounding box of the given object. On these points, Euclidean cluster extraction \cite{mci/Rusu2010} is used, as they can belong to multiple objects. Also, ground segmentation is done for the LIDAR measurements by M-estimator SAmple Consensus (MSAC) plane fitting \cite{TORR2000138} in order to filter ground points from the frustum (Fig. \ref{fig:ground}).

Now, a point-level tracking is implemented from $I_{t-1}$ to $I_{t}$ using the Kanade-Lucas-Tomasi (KLT) algorithm \cite{5596017}.

\begin{figure}[!h]
  \vspace{-0.2cm}
  \centering
  {\includegraphics[width=3in]{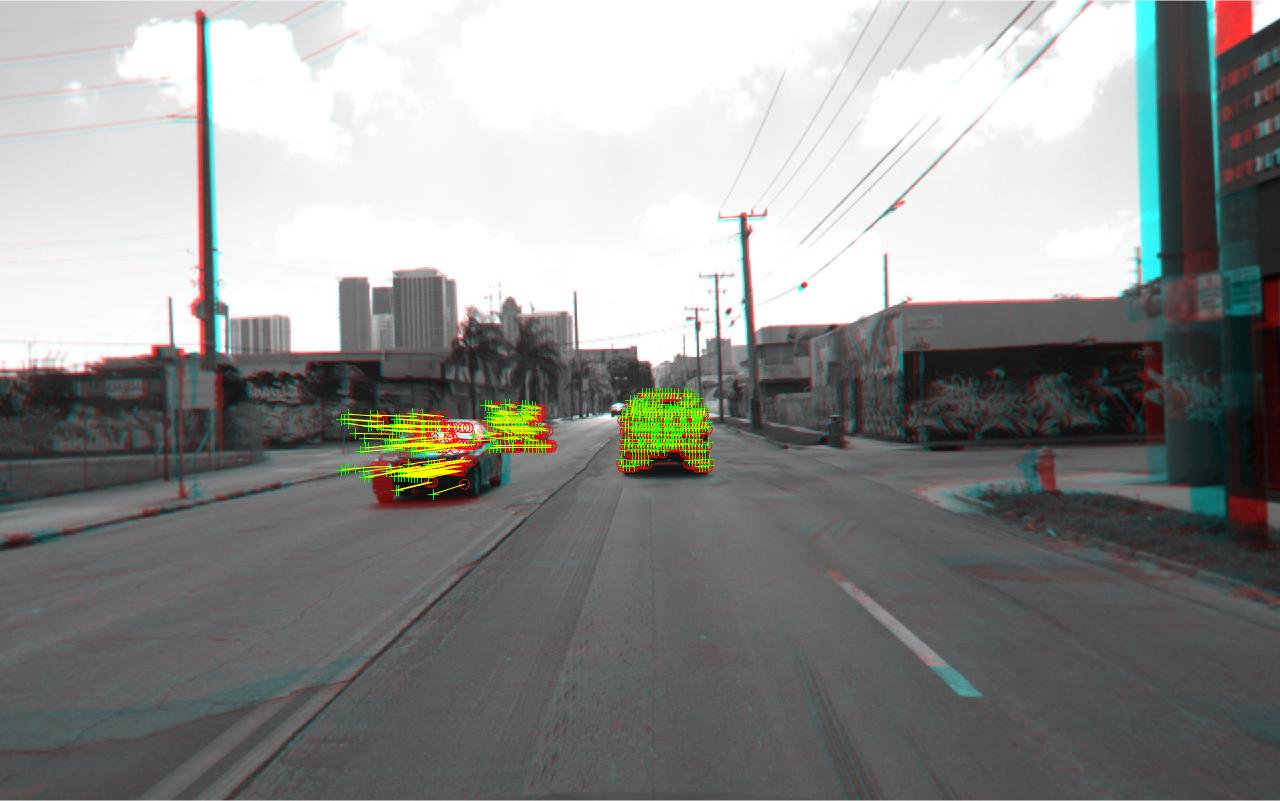}}
  \caption{Projected objects points (from LIDAR) tracked from $I_{t-1}$ to $I_{t}$. The images are overlayed.}
  \label{fig:pointtrack}
  \vspace{-0.1cm}
\end{figure}

Illustration of 2D points projected from the LIDAR and tracked to the next frame for each object can be seen in Fig. \ref{fig:pointtrack}.

\subsection{Estimate Virtual Camera Pose}
\label{subsec:virpos}
As it was said earlier, we will use the tracked object points on $I_{t}$ to estimate object transformations. To do that, we need to estimate the virtual camera pose assuming static objects and compensate the estimations with the ego-motion. (We used ground truth poses provided by the different datasets in the experiments.) %In our experiments we used the GPS-based poses provided by the different datasets.
Now, the virtual camera pose is estimated based on the 3D ($P_{t-1}$) - 2D ($I_{t}$) correspondences. We estimate for each $i^{th}$ dynamic object the $\mathbf{T}_{t,i}$ transformation matrix, mapping the object points from the global coordinate system (coordinate system of the previous LIDAR frame $P_{t-1}$) to the image coordinates of the actual camera frame ($I_{t}$):

\begin{equation}
\begin{bmatrix}
u_{t} \\
v_{t} \\
1
\end{bmatrix}
=
\mathbf{K} \cdot \mathbf{T_{t,i}}  \cdot
\begin{bmatrix}
X_{t-1} \\
Y_{t-1} \\
Z_{t-1} \\
1
\end{bmatrix}
\end{equation}

%In our experiments point pairs with at least 50 matches were evaluated. 
The Perspective-n-Point problem has been solved by \cite{1217599} with MLESAC \cite{Torr00}. According to our experiments, this model fitting is robust against outlier points (which may have remained from the previous steps). In Section \ref{subsec:err}, there is a discussion about how the number of object points affects the estimation accuracy.

%The transformations from the previous LIDAR frame, $P_{t-1}$ to the virtual frame of $P_{t,v}$ is calcuated as:

If we are interested in the movement of the given object (e.g. for calculating velocity), we can express the transformation matrix, $\mathbf{T}_{mov,i}$ (in the coordinate system of $P_{t-1}$). This is done based on the equality of the coordinate systems of the virtual camera position and the real camera position at $t$ timestamp. The movement of each object is given by:

%\begin{equation}
%    \mathbf{T}_{t} \cdot \mathbf{T}_{mov,i} \cdot \mathbf{T}_{t-1}^{-1} = \mathbf{T}_{t,i} \cdot \mathbf{T}_{t-1,i}^{-1}
%\end{equation}

%$\mathbf{T}_{t-1}$ and $\mathbf{T}_{t-1}$ transforms from the global coordinate system to the coordinate systems of the real camera poses at $t-1$ and $t$ respectively. (Choosing the coordinate system of the first LIDAR frame as the global coordiante system $\mathbf{T}_{t-1} = T_{L,C}^{-1}$.) These matrices assumed to be known, as the ego-motion is known. $\mathbf{T}_{t-1,i}$ and $\mathbf{T}_{t,i}$ transforms from the global coordinate system to the coordinate systems of the virtual camera poses at $t-1$ and $t$ respectively ($\mathbf{T}_{t-1,i} = T_{L,C}^{-1}$) defined for each objects separetely.
%The equation above describes that the real and virtual camera positions have the same coordinate systems both caseos of $t-1$ and $t$ timestamps. The movement of each object is given by:

%\begin{equation}
%    \mathbf{T}_{mov,i} \cdot \mathbf{T}_{t-1}^{-1} = \cdot \mathbf{T}_{t-1,i}^{-1}
%\end{equation}

\begin{equation}
    \mathbf{T}_{mov,i} = \mathbf{T}_{t}^{-1} \cdot \mathbf{T}_{t,i} 
\end{equation}

where $\mathbf{T}_{t}$ indicates the transformation from the global coordinate system to the coordinate systems of the real camera poses at $t$. This is assumed to be known, as the ego-motion is known.

\subsection{Generating $P_{t,v}$, Virtual LIDAR Cloud at Timestamp $t$}
To generate the virtual point cloud at timestamp $t$ two transformations are necessary.
First, we transform the 3D points of the static objects from the coordinate system of the previous LIDAR frame to the actual (virtual) one by $T_{S}$:

\begin{equation}
    \mathbf{T}_{S}=\mathbf{T}_{L,C}^{-1} \cdot \mathbf{T}_{t}
\end{equation}

%\begin{equation}
%    \mathbf{T}_{S}=\mathbf{T}_{L,C}^{-1} \cdot \mathbf{T}_{t} \cdot \mathbf{T}_{t-1}^{-1} \cdot \mathbf{T}_{L,C}
%\end{equation}

Then, we express ${T}_{D,i}$ from ${T}_{t,i}$-s and iterate through each $i$ dynamic object candidates to transform their points between LIDAR coordinate frames:

%\begin{equation}
%    \mathbf{T}_{D,i}=\mathbf{T}_{L,C}^{-1} \cdot \mathbf{T}_{t} \cdot \mathbf{T}_{mov,i} \cdot \mathbf{T}_{t-1}^{-1} \cdot \mathbf{T}_{L,C}
%\end{equation}

\begin{equation}
\mathbf{T}_{D,i} = \mathbf{T}_{L,C}^{-1} \cdot \mathbf{T}_{t,i}
\end{equation}

\begin{figure*}[h]
\centering
\subcaptionbox{Point cloud of the last frame; vehicles are colored with blue. \label{fig:t_1}}
{\includegraphics[width=0.51\textwidth]{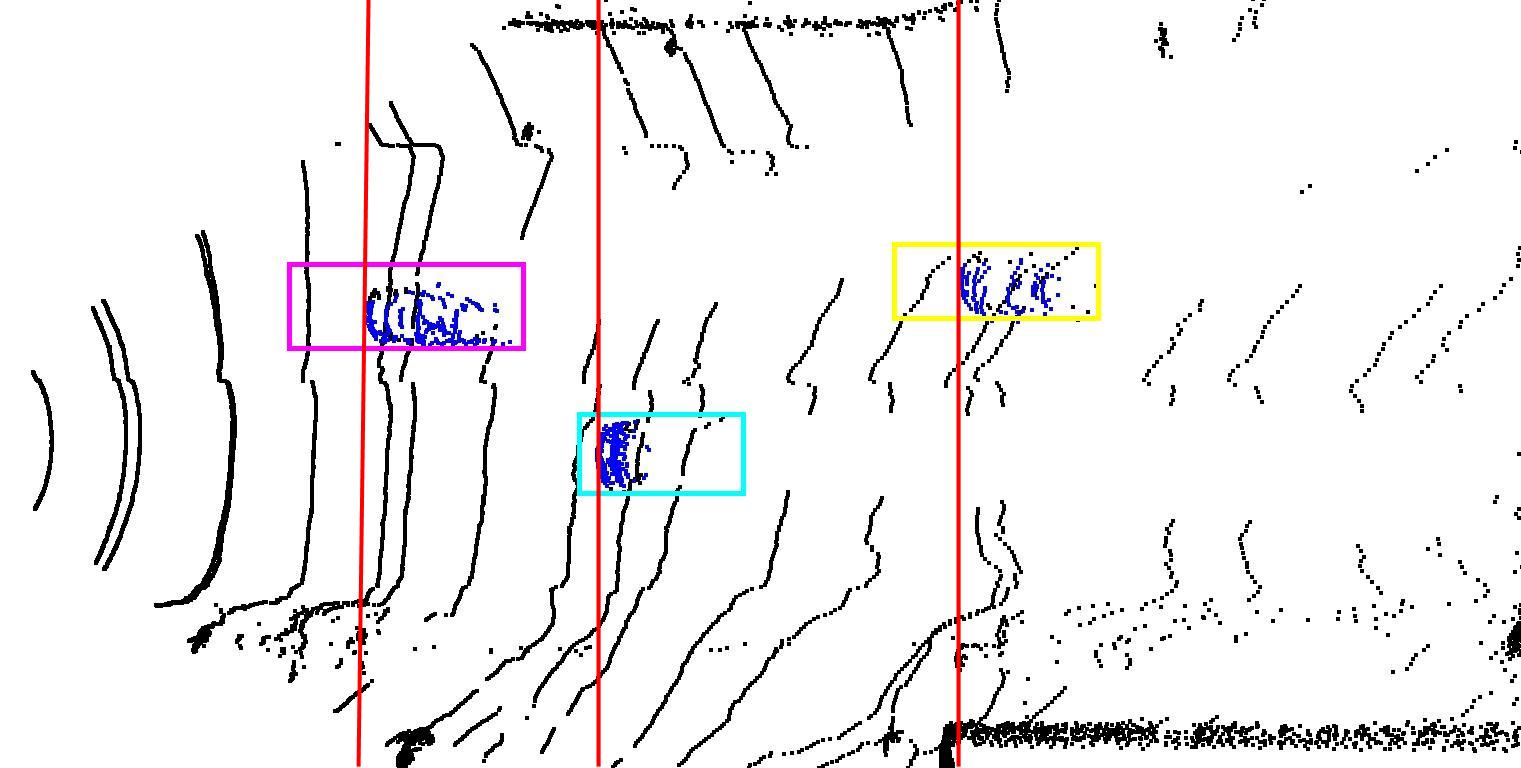}}
\subcaptionbox{Our estimation to the time instance of the next intermediate camera frame; vehicles are colored with green. \label{fig:t_next}}
{\includegraphics[width=0.49\textwidth]{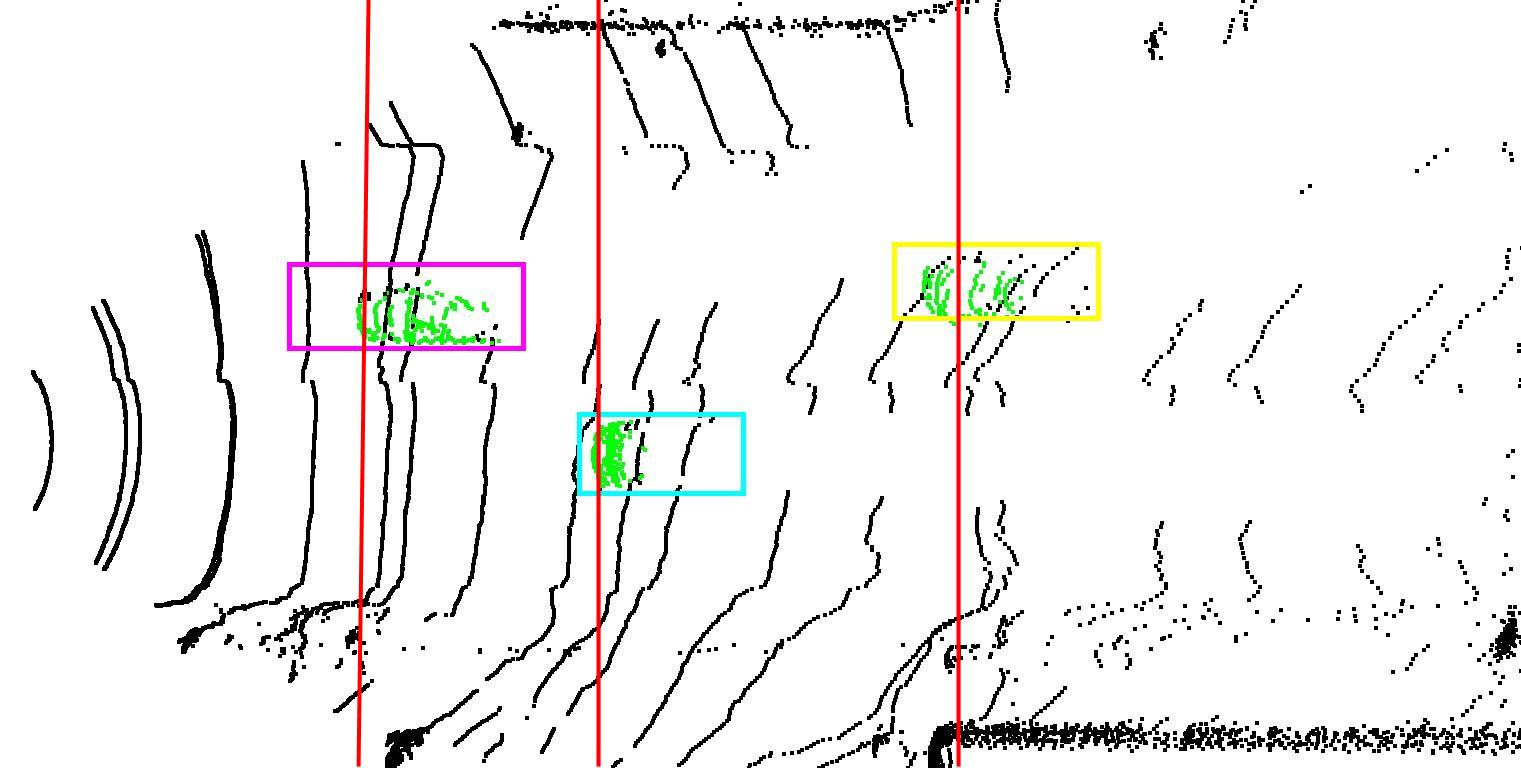}}
\subcaptionbox{Our estimation to the time instance of intermediate camera frame after the next; vehicles are colored with green. \label{fig:t_nex2}}
{\includegraphics[width=0.49\textwidth]{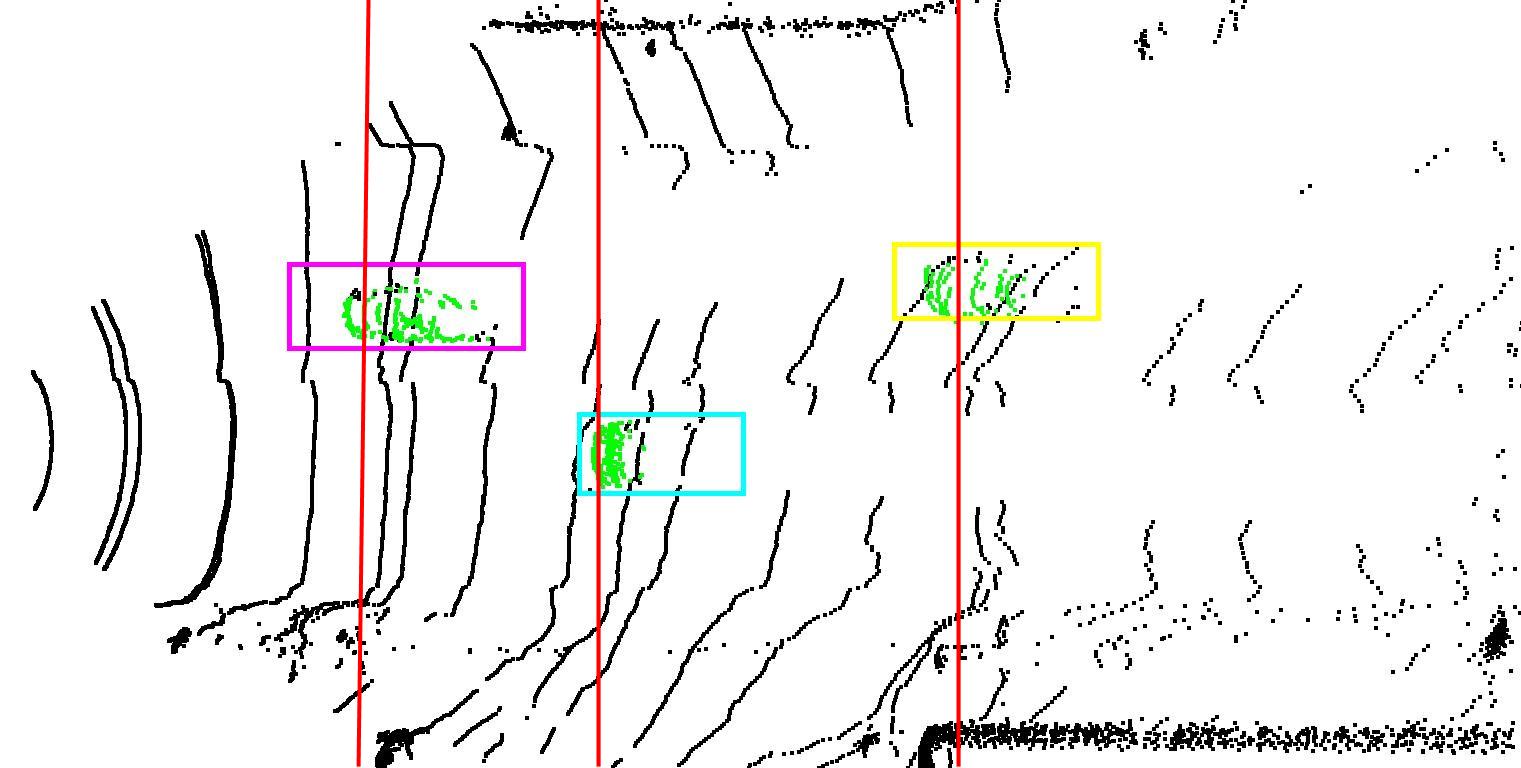}}
\subcaptionbox{Next available LIDAR frame (ground truth); vehicles are colored with red. \label{fig:t_gt}}
{\includegraphics[width=0.49\textwidth]{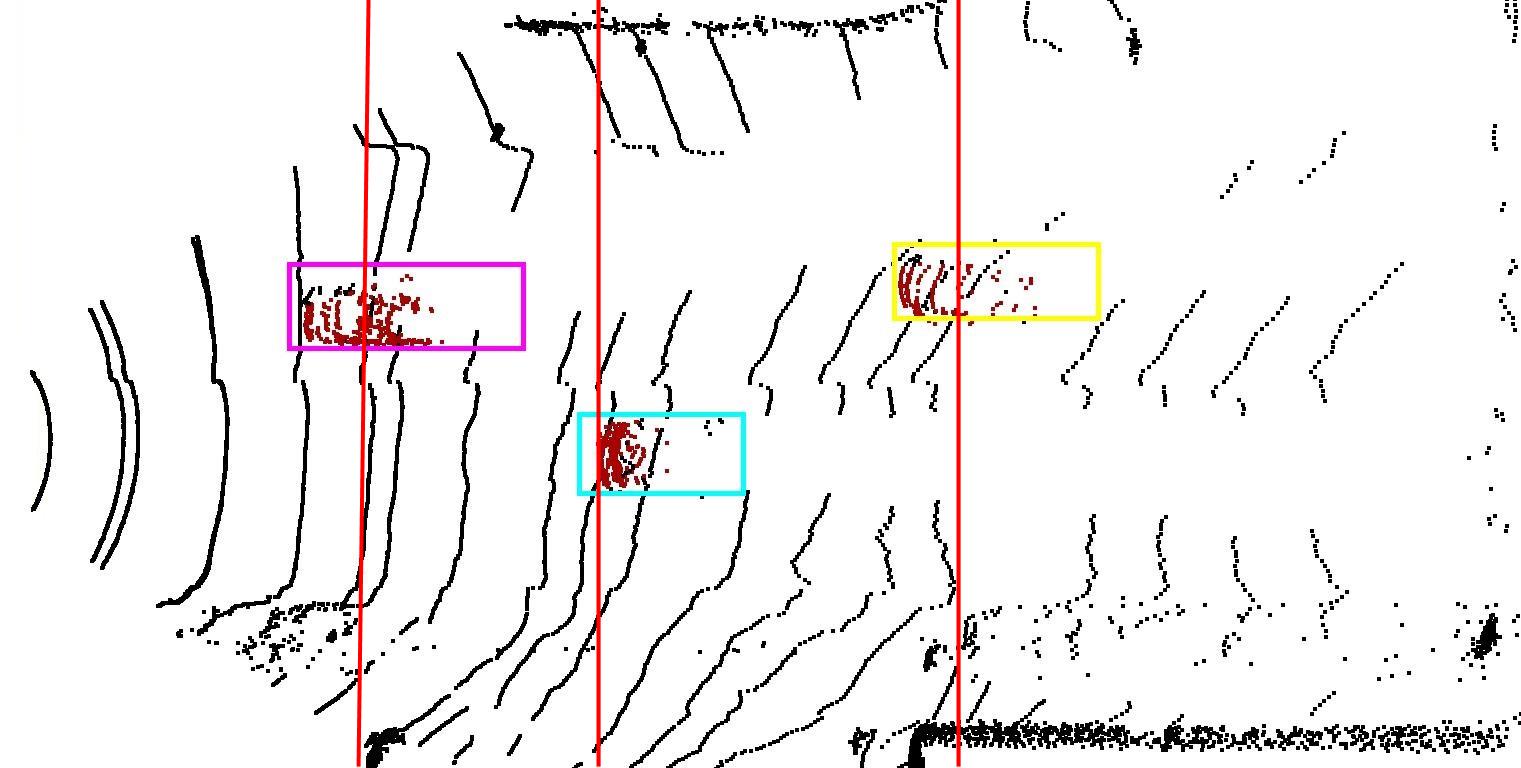}}
\subcaptionbox{Our estimation to the time instance of the next available LIDAR frame (ground truth); vehicles are colored with green. \label{fig:t_e}}
{\includegraphics[width=0.49\textwidth]{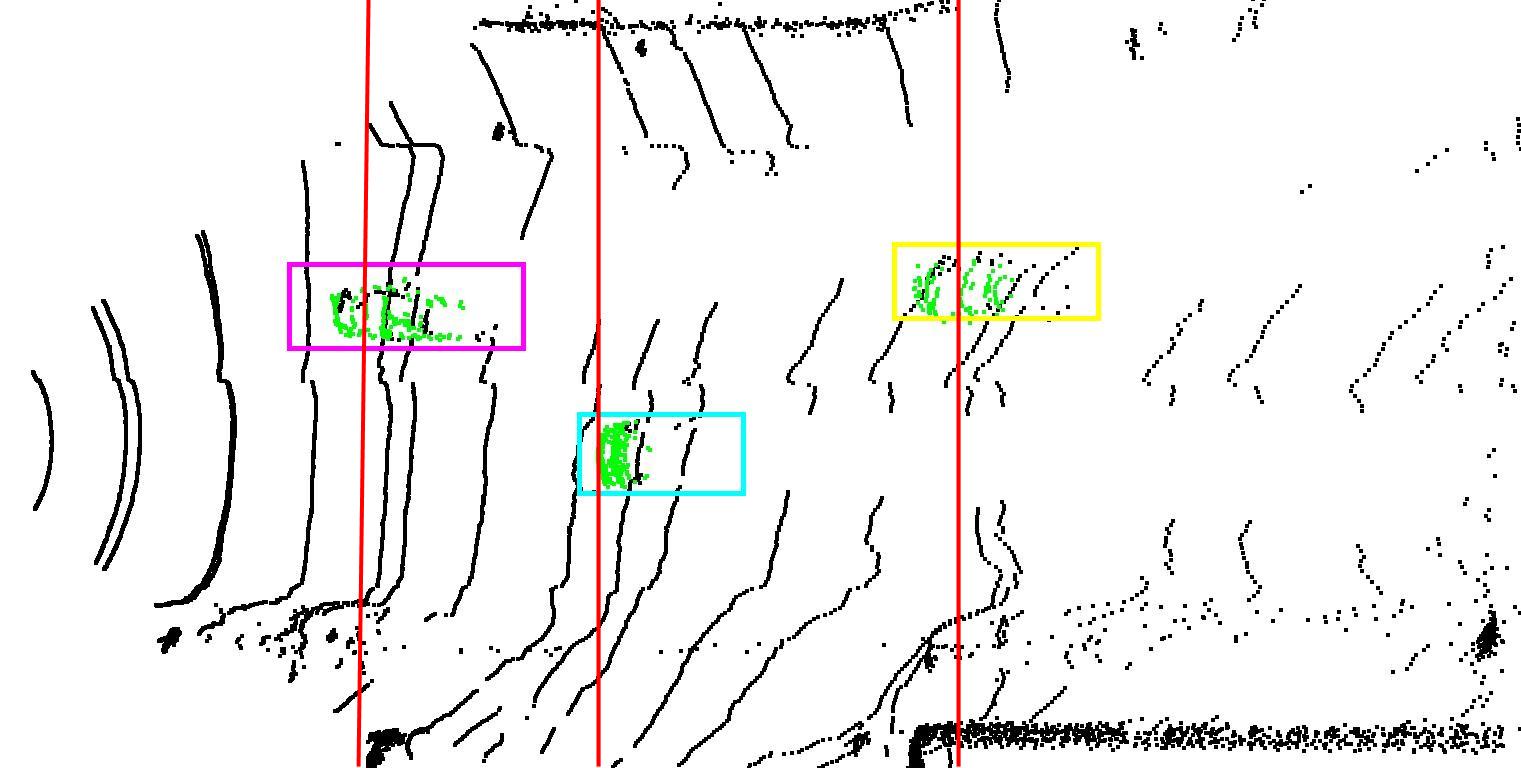}}
\subcaptionbox{Last measurement of yellow highlighted vehicle (blue) and its ground truth position (red) \label{fig:gt_t}}
{\includegraphics[width=0.3\textwidth]{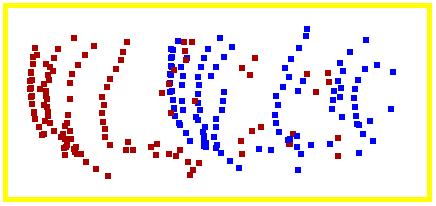}}
\subcaptionbox{Our estimation of the yellow highlighted vehicle (green) and its ground truth position (red)  \label{fig:gt_es_t}}
{\includegraphics[width=0.3\textwidth]{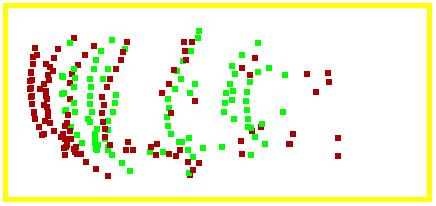}}
\caption{Illustration of the proposed method. (The point clouds corresponds to the images of Fig. \ref{fig:track}, the bounding boxes of the same vehicles have the same colors.) (a) subfigure shows the last available LIDAR frame, (b) and (c) shows our LIDAR frame estimation using intermediate camera frames, (d) shows the next available LIDAR measurement (ground truth), while (e) our estimation to this timestamp generated with our method. (Red vertical lines indicate the positions vehicles at $t-1$.)
Subfigures (f) highlights ground truth points of the vehicle with yellow bounding box at two consecutive LIDAR frames. While (g) visualizes together the ground truth point cloud and our estimation to the given time instance.
}\label{fig:pc_est}
\end{figure*}

In this way, we have generated $P_{t,v}$, our virtual LIDAR point cloud at timestamp $t$ (where we originally had no LIDAR measurement), transforming the points of the previous measurement based on our geometric estimation. We can upsample our LIDAR measurements temporally by repeating these steps for any $t-1$ and $t$. Sample result are illustrated on Fig. \ref{fig:pc_est}.

%\begin{figure}[!h]
%\centering
%\subcaptionbox{Point cloud of $T_{t-1}$ (detected dynamic objects are colored with blue) \label{fig:t_1}}
%{\includegraphics[width=0.51\columnwidth]{Figures/frame_t_1_2.png}}
%\subcaptionbox{Ground truth point cloud of $T_{t}$ (detected dynamic objects are colored with red) \label{fig:gt}}
%{\includegraphics[width=0.49\columnwidth]{Figures/frame_t_2.png}}
%\subcaptionbox{Estimated point cloud of $T_{t}$ (dynamic objects are colored with green) \label{fig:es}}
%{\includegraphics[width=0.49\columnwidth]{Figures/frame_t_es_2.png}}
%\subcaptionbox{Yellow highlighted vehicle at $T_{t-1}$ (blue) and $T_{t}$ (red) \label{fig:gt_t}}
%{\includegraphics[width=0.49\columnwidth]{Figures/t_1_gt_t_gt.png}}
%\subcaptionbox{Yellow highlighted vehicle's ground truth (red) and estimated (green) position at $T_{t}$ %\label{fig:gt_es_t}}
%{\includegraphics[width=0.49\columnwidth]{Figures/t_es_gt.png}}
%\caption{Illustration of point cloud estimation of the proposed method. (a) subfigure shows the last LIDAR frame %($P_{t-1}$), (b) shows the ground truth and (c) our estimation of the point cloud at $T_{t}$. 
%Subfigure (d) highlights ground truth points of a vehicle at $T_{t-1}$ and $T_{t}$. While (e) visualizes together the ground truth point cloud and our estimation.}\label{fig:pc_est}
%\end{figure}

\section{Results}
\label{sec:test}
There is no method available in the literature designed to solve the temporal LIDAR upsampling problem. Here, we compare our method to ones that can be alternatives. These methods are very distinct and so use different datasets for evaluation. We also evaluated using these datasets for a fair comparison. These are the Argoverse dataset \cite{DBLP:journals/corr/abs-1911-02620}, and different variants of the KITTI dataset \cite{Geiger2013IJRR}. As in most of the data only front facing cameras were available, we generated point clouds in this common field of view. If there were available ground truth detection, we used them to evaluate independently from the detector performance. As the detector we use, described in subsection \ref{det_and_track} is a preprocessing step for our point of view and also replaceable.
The commonly applied error metrics for point cloud generation are Chamfer distance (CD) and Earth Movers Distance (EMD) \cite{Fan_2017_CVPR}. 

The formula for chamfer distance:

\begin{equation}
\begin{split}
    CD = \frac{1}{N} \sum_{p_{t,v} \in P_{t,v}} \min_{p_{t} \in P_{t}} \Vert p_{t}-p_{t,v} \Vert^{2} \\ + \quad \frac{1}{N} \sum_{p_{t} \in P_{t}} \min_{p_{t,v} \in P_{t,v}} \Vert p_{t}-p_{t,v} \Vert^{2}
    \end{split}
    \label{chamfer_def}
\end{equation}

where $p_{t,v} \in \mathbb{R}^{N_{t,v} \times 3}$  and $p_{t} \in \mathbb{R}^{N_{t} \times 3}$ are the data points of the predicted (virtual) $P_{t,v}$  and ground truth $P_{t}$ point clouds respectively. In our case the number of generated (transformed) points ($N_{t,v}$) can differ from the number of data points of the ground truth $N_{t}$ in case of object-level evaluation. That is why the denser point cloud is randomly downsampled to $N$ for these scenarios, where $N= \min({N_{t,v},N_{t}}$).

The Earth Movers Distance:

\begin{equation}
    EMD = \frac{1}{N} \sum_{p_{t} \in P_{t}} \Vert p_{t}- \phi (p_{t,v}) \Vert^{2}
     \label{emd_def}
\end{equation}

where $\phi$ is bijection, which calculates the point-to-point mapping between two point cloud $P_{t}$ and $P_{t,v}$.

EMD calculates the cost of the global matching problem; in this way, it measures similarity between point clouds. CD is a distance measure between the nearest neighbors of points back-and-forth. We calculated EMD with the approximation of \cite{4048607} used by \cite{Fan_2017_CVPR} and other literature giving reference. %Unfortunately, different definitions of the above metric are used in the literature. These vary in using euclidean distance or squared Euclidean distance in both metrics. Also, the consistency between their reported values and published definitions is questionable (see our results and reported results in the result Tables). That is why we report our results in both cases (euclidean distances and squared euclidean distances). The applied definition is indicated in the unit of the metric.

To evaluate the point cloud generation capability of our pipeline, we generated virtual clouds at time instances where real measurements were available and used them as ground truth. We evaluated frames containing dynamic object candidates.

For the evaluation of the generated virtual LIDAR measurement, we applied the same postprocessing as the authors of a given proposal we compare to. In most of the cases it was the proposition suggested by \cite{9652466}, \cite{Lu2020_PointINet} and \cite{spinet} (if not it is indicated). Namely, we also downsampled the LIDAR frames to 16384 data points. After that, we only considered points for which our estimation is applied, points seen by the specific camera used for upsampling (in case of Argoverse dataset front facing center camera, in case of KITTI dataset left RGB camera).

\subsection{Argoverse Dataset}
Argoverse is frequently used for evaluation in the point cloud forecasting domain. 
We used the validation sequences of the Argoverse dataset for the evaluation, containing 5015 LIDAR frames. 

%\begin{table}[h]
%\caption{Quantitative evaluation of the proposed pipeline on Argoverse dataset.}\label{tab:results} \centering
%\begin{tabular}{|c|P{1cm}|P{1cm}|c|}
%\hline
%  Methods &  MoNet (LSTM) \cite{9652466} & MoNet (GRU) \cite{9652466} & Proposed\\\hline
%  CD [$m$]  & 2.105 & 2.069 & \textbf{0.8109}\\\hline
%  CD [$m^{2}$]  & - & - &  1.9898\\\hline
%  EMD [$m$] & 368.21 &  364.14 & \textbf{3.2456}\\\hline
%  EMD [$m^{2}$] & - & - & 208.3344\\
%  \hline
%\end{tabular}
%\end{table}

%\begin{table}[h]
%\caption{Quantitative evaluation of the proposed pipeline on Argoverse dataset.}\label{tab:results} \centering
%\begin{tabular}{|P{1.5cm}|P{0.9cm}|P{0.9cm}|P{0.9cm}|P{0.9cm}|}
%\hline
%  Methods & CD [$m$] & CD [$m^{2}$] & EMD [$m$] & EMD [$m^{2}$]\\\hline
%  MoNet (LSTM) \cite{9652466} & 2.105 & - & 368.21 & - \\\hline
%  MoNet (GRU) \cite{9652466} & 2.069 & - & 364.14 & - \\\hline
%  % whole frame part (just about 7-8) sequence were run for this
%  %Proposed & \textbf{0.64} & &\textbf{4.07} &\\
%  % object framenként
%  %Proposed & \textbf{0.465} & 0.6938 & \textbf{1.1417} & 15.9260 
%  Proposed method & \textbf{0.811} &  1.990 & \textbf{3.25} & 208.33\\ %& 
%  \hline
%\end{tabular}
%\end{table}

\begin{table}[h]
\caption{Quantitative evaluation of the proposed pipeline on Argoverse dataset.}\label{tab:results} \centering
\begin{tabular}{|P{3.0cm}|P{1.5cm}|P{1.5cm}|}
\hline
  Methods & CD [$m^{2}$] & EMD [$m^{2}$]\\\hline
  MoNet (LSTM) \cite{9652466} & 2.105 &  368.21  \\\hline
  MoNet (GRU) \cite{9652466} & 2.069  & 364.14  \\\hline
  \textbf{Proposed method} &   \textbf{1.990} &  \textbf{208.33}\\ %& 
  \hline
\end{tabular}
\end{table}

Table \ref{tab:results} shows that we are superior to other methods on the Argoverse dataset in the case of virtual frame generation.
 \cite{9389722} also used the Argoverse dataset for the evaluation of their prediction capability. However, they only predicted point clouds in the near range of ({\it 10m $\times$ 10m}). %That is why it is not directly comparable to methods generating a whole LIDAR frame, but 
 In \cite{ICIAP_ZR_TSZ} one can find a comparison to our earlier work outperforming this method.
 
\cite{ijcv} noticed that dynamic objects have little attention in the point cloud generation domain. The main focus of our paper is the pose estimation and reconstruction of these. That is why we provide separate evaluations for vehicles in Table \ref{tab:resultsdyn}. Here, we used the postprocessing proposed by \cite{ijcv}, the point cloud is evaluated without ground, it is cropped $\left[ -32,32 \right]  \left[ -8,8 \right] \left[ -\infty, 2 \right]$ m in $x,y$ and $z$ direction and downsampled to 2048 points. As \cite{ijcv} states, the EMD is biased by outliers (especially in the case of separated objects present in the scene), so we provided results without these as well.
 
% \begin{table}[h]
%\caption{Quantitative evaluation of the proposed pipeline on Argoverse dataset (just vehicles).}\label{tab:resultsdyn} \centering
%\begin{tabular}{|P{1.5cm}|P{0.9cm}|P{0.9cm}|P{0.9cm}|P{0.9cm}|}
%\hline
%  Methods & CD [$m$] & CD [$m^{2}$] & EMD [$m$] & EMD [$m^{2}$]\\\hline
%%  PointNet++ and LSTM \cite{10.5555/3295222.3295263} & - & 1.232 &  - & 2.32 %\\\hline
%  SPCM-Net \cite{ijcv} & - & 1.345 &  - & 2.30 \\\hline
%  Proposed method (w/ outliers) & 0.322 & 0.814 & 0.61 & 4.43 \\\hline
%  Proposed method (w/o outliers) & 0.317 & \textbf{0.801} & 0.35 & %\textbf{0.94} \\\hline
%\end{tabular}
%\end{table}

\begin{table}[h]
\caption{Quantitative evaluation of the proposed pipeline on Argoverse dataset (just vehicles).}\label{tab:resultsdyn} \centering
\begin{tabular}{|P{3.0cm}|P{1.5cm}|P{1.5cm}|}
\hline
  Methods &  CD [$m^{2}$] & EMD [$m^{2}$]\\\hline
%  PointNet++ and LSTM \cite{10.5555/3295222.3295263} & - & 1.232 &  - & 2.32 \\\hline
  SPCM-Net \cite{ijcv}  & 1.345  & 2.30 \\\hline
  \textbf{Proposed method} (w/ outliers)  & 0.814  & 4.43 \\\hline
  \textbf{Proposed method} (w/o outliers)  & \textbf{0.801} &  \textbf{0.94} \\\hline
\end{tabular}
\end{table}

\subsection{KITTI Dataset}
\subsubsection{Odometry Dataset}
The odometry dataset is the most commonly used in the case of sequential point cloud forecasting methods; Point cloud interpolation networks also use this dataset for evaluation. However, the number of sequences used for evaluation is different. While \cite{9652466} used the sequences 08 to 10 for testing, \cite{Lu2020_PointINet} and \cite{spinet} executed their tests on sequences 02 to 10. The latter one contains 17500 frames. % As we do not need training data for our geometry-based method, we used sequences 02 to 10 as well.
%We used the sequences 08 to 10 for testing as \cite{9652466} did.
We evaluated both cases of 3 and 10 sequences for a fair comparison. As there was only a negligible difference in the error metrics of these cases, we reported our larger errors in Table \ref{tab:resultskito}.

\begin{table}[h]
\caption{Quantitative evaluation of the proposed pipeline on KITTI odometry dataset.}\label{tab:resultskito} \centering
\begin{tabular}{|P{3.0cm}|P{1.5cm}|P{1.5cm}|}
\hline
  Methods  & CD [$m^{2}$]  & EMD [$m^{2}$]\\\hline
  MoNet (LSTM) \cite{9652466} & 0.573  & 91.79  \\\hline
  MoNet (GRU) \cite{9652466} & 0.554  & 91.97 \\\hline
  SPINet \cite{spinet} & 0.465  & 40.69 \\\hline
  PointINet \cite{Lu2020_PointINet} & \textbf{0.457}  & 39.46 \\\hline
  \textbf{Proposed method} &  0.471  & \textbf{33.98}\\% 
 
  \hline
\end{tabular}
\end{table}

%\begin{table}[h]
%\caption{Quantitative evaluation of the proposed pipeline on KITTI odometry dataset (compared to offline interpolation).}\label{tab:results} \centering
%\begin{tabular}{|c|c|c|}
%\hline
%  Methods & CD [$m$] & EMD [$m$]\\\hline
%  \cite{Lu2020_PointINet} & 0.457 & 39.46 \\\hline
%  Proposed & \textbf{0.3830} & \textbf{1.78}\\
%  \hline
%\end{tabular}
%\end{table}

%\begin{table}[h]
%\caption{Quantitative evaluation of the proposed pipeline on KITTI odometry dataset (just vehicles).}\label{tab:resultskito2} \centering
%\begin{tabular}{|c|P{1cm}|P{1cm}|c|}
%\hline
%  Methods & \cite{Weng2020_SPF2} point based & \cite{Weng2020_SPF2} range map based & Proposed\\\hline
%  CD [$m$]  & - & - & 0.5379\\\hline
%  CD [$m^{2}$]  & 2.37 & 0.92 & \textbf{0.63}\\\hline
%  EMD [$m$] & - &  - & 1.8763\\\hline
%  EMD [$m^{2}$] & 211.47 & 128.81 & \textbf{31.03}\\
%  \hline
%\end{tabular}
%\end{table}

%\begin{table}[h]
%\caption{Quantitative evaluation of the proposed pipeline on KITTI odometry dataset (just vehicles).}\label{tab:resultskito2} \centering
%\begin{tabular}{|P{1.5cm}|P{0.9cm}|P{0.9cm}|P{0.9cm}|P{0.9cm}|}
%\hline
%  Methods & CD [$m$] & CD [$m^{2}$] & EMD [$m$] & EMD [$m^{2}$]\\\hline
%  \cite{Weng2020_SPF2} point based & - & 2.37 & - & 211.47 \\\hline
%  \cite{Weng2020_SPF2} range map based & - & 0.92 & - & 128.81 \\\hline
%  Proposed method (w/ outliers) & 0.54  & 0.63 & 1.88 & 31.03\\ % 
%  \hline
%  Proposed method (w/o outliers) & 0.4730  & \textbf{0.4231} & 0.527 & %\textbf{1.91}\\ % 
%  %KITTI odometry dataset & 0.397 & 1.114 &  0.468 & 1.63 \\\hline
%  \hline
%\end{tabular}
%\end{table}

\begin{table}[h]
\caption{Quantitative evaluation of the proposed pipeline on KITTI odometry dataset (just vehicles).}\label{tab:resultskito2} \centering
\begin{tabular}{|P{3.0cm}|P{1.5cm}|P{1.5cm}|}
\hline
  Methods & CD [$m^{2}$]  & EMD [$m^{2}$]\\\hline
  \cite{Weng2020_SPF2} point based  & 2.37  & 211.47 \\\hline
  \cite{Weng2020_SPF2} range map based  & 0.92  & 128.81 \\\hline
  \textbf{Proposed method} (w/ outliers)   & 0.63  & 31.03\\ % 
  \hline
  \textbf{Proposed method} (w/o outliers)  & \textbf{0.42} & \textbf{1.91}\\ % 
  \hline
\end{tabular}
\end{table}

%\begin{table}[h]
%\caption{Quantitative evaluation of the proposed pipeline on KITTI odometry %dataset (just vehicles).}\label{tab:resultskito2} \centering
%\begin{tabular}{|c|c|c|}
%\hline
%  Methods & CD [$m^{2}$] & EMD [$m^{2}$]\\\hline
%  \cite{Weng2020_SPF2} point based & 2.37 & 211.47 \\\hline
%  \cite{Weng2020_SPF2} range map based & 0.92 & 128.81 \\\hline
%  Proposed & \textbf{0.63} & \textbf{31.03}\\ % & 0.5379 & 1.8763
%  \hline
%\end{tabular}
%\end{table}

Table \ref{tab:resultskito} shows our results compared to point cloud forecasting methods of \cite{9652466} and also to point cloud interpolation method of \cite{Lu2020_PointINet} and \cite{spinet}. The last ones have the most similar purpose to our method. Our CD value is somewhat below compared to them. However, EMD is better to characterize the similarity of point clouds. Also, they can operate only offline as they utilize 'future' frames for the interpolation. These results prove that our pipeline is most applicable for generating virtual LIDAR frames (to time instances when there were no available) in real scenarios.
 
 The main idea of our method is the localization of dynamic objects at these instances. That is why we provide an evaluation separately for these (just in case of Argoverse dataset) at Table \ref{tab:resultskito2} compared with the performance of a method to the similar task \cite{Weng2020_SPF2}.
  We outperformed \cite{Weng2020_SPF2} by a large margin as well. Also, a comparison to methods designed for localization of surrounding vehicles is provided in the following subsection.

 %As the transformation of the static scene was known, for fair evaluation, we only evaluated our prediction accuracy for the points of dynamic object candidates.
  %Previous results on its test set was published in \cite{ICIAP_ZR_TSZ}.

\subsubsection{Trajectories from the RAW Dataset}
As we said in Section \ref{sec:relw}, some methods are not capable of generating point clouds (with LIDAR characteristics) but can localize surrounding vehicles in 3D (in time instances where LIDAR measurements are not available). In this way, they can be an alternative to our proposal of Section \ref{subsec:virpos}. These methods are usually evaluated at specific sequences of the KITTI raw dataset (more than 1200 frames) on specific vehicle trajectories and use depth error as a performance measure. So, we evaluated our proposal on this data with the given performance measure as well. Results are shown in Table \ref{real_kitti}.

\begin{table*}
\caption{Results of dynamic object localization from KITTI raw dataset}
\begin{center}
\begin{tabular}{|c|c||c|c|c|c||c|c|c|c||c||c|}
\hline
      \multicolumn{2}{|c||}{Sequence id } & \multicolumn{4}{c||}{04}  & \multicolumn{4}{c||}{47} & 56 &\multirow{2}{*}{Average}  \\\cline{1-11}
     \multicolumn{2}{|c||}{Object id } & 1 & 2 & 3 & 6 & 0 & 4 & 9 & 12 & 0 & \\\hline
      \multirow{3}{*}{Depth error [\%]}   & \cite{7298997} & 6.0 & 5.6 & 4.9 & 5.9 & \textbf{5.9} & 12.5 & 7.0 & 8.2 & 6.0 & 6.8\\\cline{2-12}
    & \cite{8708251} & 5.1 & 11.1 & 1.8 & 2.5 & 9.3 & 3.8 & 6.6 & 2.8 & \textbf{0.8} & 4.9\\\cline{2-12}
     & proposed & \textbf{1.3} & \textbf{1.1} & \textbf{0.9} & \textbf{1.5} & 6.2 & \textbf{1.7} & \textbf{0.6} & \textbf{1.6} & \textbf{0.8} & \textbf{1.7}\\\hline
\end{tabular}
\label{real_kitti}
\end{center}
\end{table*}

As expected, we significantly outperformed the methods that use only one camera information. So, they are not alternatives for the upsampling problem in terms of accuracy.

\section{Discussion}
\label{sec:disc}
This section provides further analysis on the proposed method, including how performance measure varies on different data and running time analysis.

\subsection{Error distribution}
\label{subsec:err}

%\begin{table}[h]
%\caption{Quantitative evaluation of the proposed estimation of dynamic point clouds.}\label{tab:resultsdyn} \centering
%\begin{tabular}{|P{1.5cm}|P{0.9cm}|P{0.9cm}|P{0.9cm}|P{0.9cm}|}
%\hline
%  Methods & CD [$m$] & CD [$m^{2}$] & EMD [$m$] & EMD [$m^{2}$]\\\hline
%  % a kommentalt resz a framenkenti osszegzes ugyanolyan mennyisegu pontokkal
%  Argoverse dataset & 0.334 & 0.2627 & 0.366 & 0.587 \\\hline
%  %Argoverse dataset & 0.317 & 0.801 & 0.35 & 0.94 \\\hline
%  KITTI odometry dataset & 0.529 & 0.664 &  0.55 & 1.49 \\\hline
%  %KITTI odometry dataset & 0.397 & 1.114 &  0.468 & 1.63 \\\hline
%\end{tabular}
%\end{table}

\begin{figure}[]
  \vspace{-0.2cm}
  \centering
  {\includegraphics[width=3in]{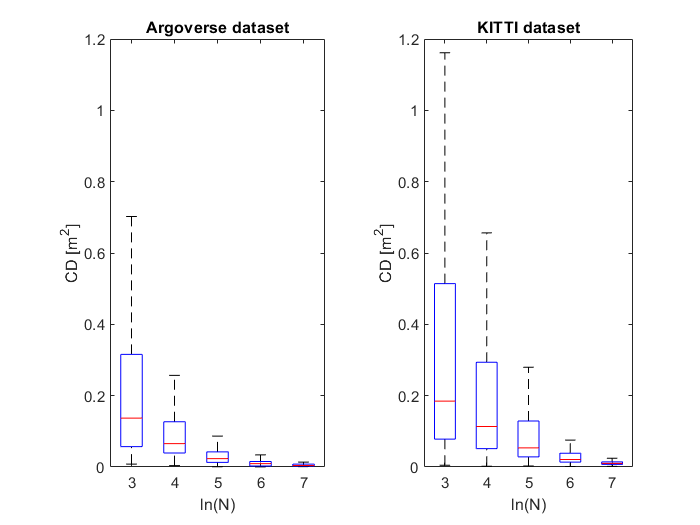}}
  \caption{Error distribution of Chamfer distance on Argoverse and KITTI odometry dataset as a function of point numbers. %Note: outliers of quartiles are not illustrated for better visualization, but they are part of the quantitative evaluation.}%; Black - static points of the first LIDAR frame.
  }
  \label{fig:errordiscd}
  \vspace{-0.1cm}
\end{figure}

\begin{figure}[]
  \vspace{-0.2cm}
  \centering
  {\includegraphics[width=3in]{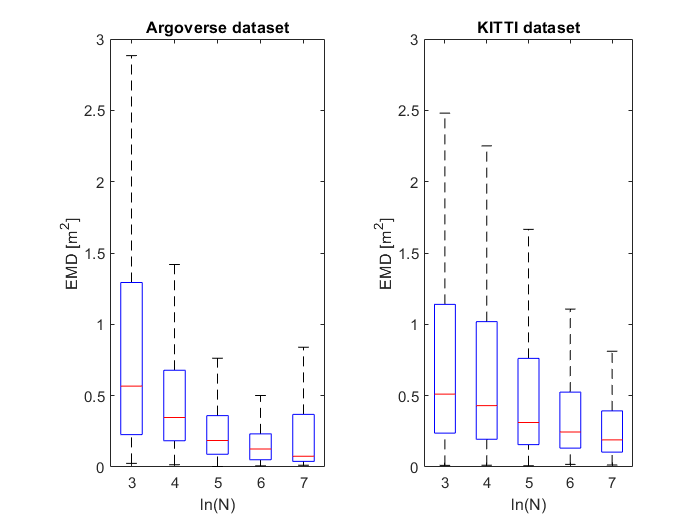}}
  \caption{Error distribution of Earth Movers distance on Argoverse and KITTI odometry dataset as a function of point numbers. %Note: outliers of quartiles are not illustrated for better visualization, but they are part of the quantitative evaluation.}%; Black - static points of the first LIDAR frame.
  }
  \label{fig:errordisemd}
  \vspace{-0.1cm}
\end{figure}

Here, we provide the distribution of the performance measures of our vehicle point cloud generation proposal. Fig. \ref{fig:errordiscd} and \ref{fig:errordisemd} illustrates the distribution of the Chamfer Distance and Earth Movers Distance (Eq. \ref{chamfer_def} and \ref{emd_def}) respectively. These figures corresponds to estimated object clouds evaluated separately in the range of $\left[ -32,32 \right]  \left[ -8,8 \right] \left[ -\infty, 2 \right]$ m in $x,y$ and $z$ (proposed by \cite{ijcv}, focusing on objects on the road). In the $x$-axis bin edges of number of data points is indicated in a logarithmic scale. Outliers of quartiles are not illustrated for better visualization, but they were part of the quantitative evaluation.
%The median values are far below the reported average and in case of EMD these are stable through point numbers and sequences. 
%Both Tables illustrates that mean of these values is not a good measure because some erroneous estimations. 
 %This statement is especially true for Chamfer distance (calculated with squared Euclidean distances). This value shows the most percent difference, if one compares values of Table \ref{tab:resultskito} to Table \ref{tab:resultskito2}. This can be explained by the fact that there are always points nearby if all the data points are considered (even if they do not correspond to the same objects). However, if only positions of specific objects are estimated, other objects' points cannot decrease CD in case of erroneous estimation, and they will increase the mean value significantly because of the squared distances. These estimations correspond to very distant objects with very few data points.
 One can see that even in the case of most erroneous estimation of vehicles (less than 20 data points, very occluded or distant vehicles %which are negligible part of the measurements
 ), the median value is far below the reported average. These figures illustrate that increasing point number (decreasing object distance) increases accuracy.
 
 %\begin{figure}[H]
 % \vspace{-0.2cm}
 % \centering
 % {\includegraphics[width=3in]{ch_dist.jpg}}
 % \caption{Error distribution of Chamfer distance on KITTI odometry dataset as a function of point numbers and sequences. Note: outliers are not illustrated for better visualization, but they are part of the quantitative evaluation.}%; Black - static points of the first LIDAR frame.}
 % \label{fig:errordiscd}
 % \vspace{-0.1cm}
%\end{figure}

 %\begin{figure}[H]
 % \vspace{-0.2cm}
 % \centering
 % {\includegraphics[width=3in]{emd.jpg}}
 % \caption{Error distribution of Earth Movers distance on KITTI odometry dataset as a function of point numbers and sequences. Note: outliers are not illustrated for better visualization, but they are part of the quantitative evaluation.}%; Black - static points of the first LIDAR frame.}
 % \label{fig:errordisemd}
 % \vspace{-0.1cm}
%\end{figure}

\subsection{Running Time Analysis}

For the running time analysis, we used the KITTI dataset and compared the speed of our method to others who have reported their ones. Results can be seen in Table \ref{run}. We downscaled the RGB images to half of their original sizes for the detection system component. (Bounding boxes are upscaled afterward, and the other estimations are done on the original image.) Note: by decreasing object points, other parts of the pipeline can also be speeded up (resulting in decreased accuracy), but we did not apply that.

\begin{table}[]
    \centering
    \begin{tabular}{P{3.0cm}|P{3.0cm}}
    Pipeline component & Average running Time [ms] \\    \hline \hline 
    Object Detector + Tracker & 27 \\\hline
    Point definition and tracker & 19\\
    Virtual camera pose estimation & 16\\
    Point cloud generation (transformation) & $\approx$ 0 \\\hline \hline
    Total & 62
    \end{tabular}
    \caption{Running time of pipeline components}
    \label{run}
\end{table}

The pipeline is implemented in the Matlab environment, configuration: 
Intel Core i7-4790K @ 4.00GHz processor, 32 GB RAM, Nvidia GTX 1080 GPU, Windows 10 64 bit. 
%Intel Core i7-8550U @ 1.80GHz processor, 8 GB RAM, Nvidia Geforce MX150 GPU, Windows 10 64 bit onboard computer. 
With the current limited research configuration, we achieved more than 15 FPS running speed, so we can temporally upscale 5 Hz LIDAR measurements. Also, one should note that the first part of the pipeline (object detection and tracking) is part of any modern environment processing pipeline. Only the remaining part (running about 30 FPS) must be executed as a further operation. In fact, these first steps can be completely spared for the generated clouds.
Table \ref{runcom} compares our running time to the reported runtimes for frame generation of the alternatives. \cite{9652466} reported 280 ms generation time for 5 frames; the table shows one-fifth of this value.

\textbf{Note}: The values of Table \ref{runcom} corresponds to 65536 points in the case of competitors. We measured running time in the case of the KITTI dataset, in which a LIDAR frame contains about double of this data point number. We executed estimations of about half of the LIDAR clouds (part in front of the camera), so overall, we generated about the same number of points.

%\begin{table}[]
%    \centering
%    \begin{tabular}{P{3.0cm}|P{3.0cm}}
%    Method & Average running Time [ms] \\    \hline \hline 
%    PointINet \cite{Lu2020_PointINet} & 726\\
%    MoNet (GRU) \cite{9652466} & \textbf{56}\\
%    Proposed method & 62 \\\hline \hline
%    \end{tabular}
%    \caption{Running time of different methods}
%    \label{runcom}
%\end{table}

\begin{table}[]
    \centering
    \begin{tabular}{P{2.5cm}|P{1.75cm}|P{1.75cm}}
    Method & Average running time [ms]  & Average running time (normalized by GFLOPS) [ms]\\    \hline \hline 
    PointINet \cite{Lu2020_PointINet} & 726 & 527\\
    MoNet (GRU) \cite{9652466} & \textbf{56} & 85\\
    Proposed method & 62 & \textbf{62} \\\hline \hline
    \end{tabular}
    \caption{Running time of different methods}
    \label{runcom}
\end{table}

The competitors used different hardware for the inference of their proposed network. That is why we provided a column beside their reported values, where we normalized their reported values by the GFLOPS (billion floating-point operations per second) ratio of the reported GPUs \footnote{www.nvidia.com} and our one. Taking into account this, we achieved the best performance in running time just in case of the accuracy of the point cloud generation. Considering the fact that the competitors must run detection and tracking steps on the generated clouds, our pipeline is much more efficient.

%(e.g. \cite{9389722} reported about 5 FPS speed with n NVIDIA RTX2080Ti GPU for their system with for only 4096 data points). We measured 105 ms average running time for our whole system without that, meaning, we can temporally upscale 5 Hz LIDAR measurements by using the current limited, research configuration.

\section{Conclusions}
\label{sec:conclusion}
The paper presented a method for temporally upsampling of LIDAR measurements based on mono camera relying on 3D and projective geometry. Our method does not require learning or specific point cloud characteristics to operate. It can generate virtual measurements in real-time. We evaluated our results on more than 20.000 frames and outperformed all the methods which could be used for similar purposes in their performance measures and databases. This comparison included different variants of different methods proposed by seven papers. We intend to investigate the localization of occluded objects as well in the future.

%\section*{Acknowledgements}
%The research was supported by the Ministry of Innovation and Technology NRDI Office within the framework of the Autonomous Systems National Laboratory Program and by the Hungarian National Science Foudation (NKFIH OTKA) No. K 139485. 

%For the corresponding footnotemark use \verb+\footnotemark[...]+

%\begin{table}[h]
%\begin{center}
%\begin{minipage}{174pt}
%\caption{Caption text}\label{tab1}%
%\begin{tabular}{@{}llll@{}}
%\toprule
%Column 1 & Column 2  & Column 3 & Column 4\\
%\midrule
%row 1    & data 1   & data 2  & data 3  \\
%row 2    & data 4   & data 5\footnotemark[1]  & data 6  \\
%row 3    & data 7   & data 8  & data 9\footnotemark[2]  \\
%\botrule
%\end{tabular}
%\footnotetext{Source: This is an example of table footnote. This is an example of table footnote.}
%\footnotetext[1]{Example for a first table footnote. This is an example of table footnote.}
%\footnotetext[2]{Example for a second table footnote. This is an example of table footnote.}
%\end{minipage}
%\end{center}
%\end{table}

\bibliographystyle{bst/sn-basic}
\bibliography{sn-bibliography}% common bib file
%% if required, the content of .bbl file can be included here once bbl is generated
%%\input sn-article.bbl

%% Default %%
%%\input sn-sample-bib.tex%

\end{document}